\documentclass[lettersize,journal]{IEEEtran}
\usepackage{amsmath,amsfonts}
\usepackage{algorithm}
\usepackage{array}
\usepackage[caption=false,font=normalsize,labelfont=sf,textfont=sf]{subfig}
\usepackage{textcomp}
\usepackage{stfloats}
\usepackage{verbatim}
\usepackage{graphicx}
\def\BibTeX{{\rm B\kern-.05em{\sc i\kern-.025em b}\kern-.08em
    T\kern-.1667em\lower.7ex\hbox{E}\kern-.125emX}}
\usepackage{balance}
\usepackage{amssymb}

\usepackage{booktabs}
\usepackage[binary-units=true]{siunitx}
\usepackage{array,multirow}
\usepackage{hyperref}

\usepackage{tablefootnote}

\usepackage[noend]{algpseudocode}
\usepackage{enumitem}
\newcommand{\figref}[1]{Figure~\ref{#1}}
\newcommand{\secref}[1]{Section~\ref{#1}}

\newcommand{\tabref}[1]{Table~\ref{#1}}

\usepackage{soul}
\usepackage[dvipsnames]{xcolor}
\usepackage{xurl}
\usepackage{xspace}

\newif\ifcomment

\commenttrue

\ifcomment
\newcommand{\hongxiang}[1]{\sethlcolor{yellow}\hl{[Hongxiang: #1]}}
\newcommand{\wayne}[1]{\sethlcolor{magenta}\hl{[Wayne: #1]}}
\newcommand{\hc}[1]{\sethlcolor{cyan}\hl{[Mark: #1]}}
\newcommand{\liam}[1]{\sethlcolor{purple}\hl{[Liam: #1]}}

\else
\newcommand{\hongxiang}[1]{}
\newcommand{\wayne}[1]{}
\newcommand{\hc}[1]{}
\newcommand{\liam}[1]{}
\newcommand{\zehuan}[1]{}
\fi

\newcolumntype{L}[1]{>{\raggedright\let\newline\\\arraybackslash\hspace{0pt}}m{#1}}
\newcolumntype{C}[1]{>{\centering\let\newline\\\arraybackslash\hspace{0pt}}m{#1}}
\newcolumntype{R}[1]{>{\raggedleft\let\newline\\\arraybackslash\hspace{0pt}}m{#1}}

\begin{document}
\title{\fontsize{23}{25}\selectfont Enhancing Dropout-based Bayesian Neural Networks \\ with Multi-Exit on FPGA}
\author{Hao (Mark) Chen,
        Liam Castelli,
        Martin Ferianc,
        Hongyu~Zhou,
        Shuanglong~Liu,\\
        Wayne Luk,
        Hongxiang Fan

\thanks{This work was supported in part by the United Kingdom EPSRC under Grant EP/L016796/1, Grant EP/N031768/1, Grant EP/P010040/1, Grant EP/V028251/1 and Grant EP/S030069/1, Maxeler, Intel, Xilinx and SGIIT. }
\thanks{H. Chen, L. Castelli, Z. Zhang and W. Luk are with the Department
of Computing, Imperial College London, London, SW7 2AZ, UK.} 
 \thanks{M. Ferianc is with the Department of Electronic and Electrical Engineering, University College London, London, WC1E 6BT, UK.}
\thanks{H. Zhou is with the School of Computer Science, University of
Sydney, Sydney, Camperdown NSW 2006, Australia.}
\thanks{S. Liu is with the School of Physics and Electronics, Hunan Normal University, Changsha 410081, China.}
\thanks{H. Fan is with Samsung AI Center, Cambridge, CB1 2JH, UK. He is also affiliated with the Department of Computer Science and Technology, University of Cambridge, CB3 0FD, UK.} 
\thanks{\IEEEauthorrefmark{1} Corresponding author: Hongxiang Fan (h.fan17@imperial.ac.uk).}
}


\maketitle

\begin{abstract}
Reliable uncertainty estimation plays a crucial role in various safety-critical applications such as medical diagnosis and autonomous driving.
In recent years, Bayesian neural networks (BayesNNs) have gained substantial research and industrial interests due to their capability to make accurate predictions with reliable uncertainty estimation.
However, the algorithmic complexity and the resulting hardware performance of BayesNNs hinder their adoption in real-life applications.
To bridge this gap, this paper proposes an algorithm and hardware co-design framework that can generate field-programmable gate array (FPGA)-based accelerators for efficient BayesNNs.
At the algorithm level, we propose novel multi-exit dropout-based BayesNNs with reduced computational and memory overheads while achieving high accuracy and quality of uncertainty estimation.
At the hardware level, this paper introduces a transformation framework that can generate FPGA-based accelerators for the proposed efficient multi-exit BayesNNs.
Several optimization techniques such as the mix of spatial and temporal mappings are introduced to reduce resource consumption and improve the overall hardware performance.
Comprehensive experiments demonstrate that our approach can achieve higher energy efficiency compared to CPU, GPU, and other state-of-the-art hardware implementations. 
To support the future development of this research, we have open-sourced our code at: \url{https://github.com/os-hxfan/MCME_FPGA_Acc.git}
\end{abstract}


\section{Introduction}

Deep neural networks (DNNs) have emerged as a cutting-edge frontier of artificial intelligence, with extensive applications in various domains ranging from computer vision~\cite{dong2021survey} to natural language processing~\cite{otter2020survey}.
However, conventional DNNs are suffering from critical limitations: they operate akin to black boxes, rendering them incapable of \textit{(a)} explaining their decisions 
and (b) estimating their uncertainty reliably when making predictions~\cite{fan2021high}. 
The lack of reliable uncertainty estimation undermines the trustworthiness of conventional DNNs, making them unsuitable candidates for safety-critical applications~\cite{leibig2017leveraging, liang2018bayesian, azevedo2020stochasticyolo} where reliable confidence and uncertainty measures are imperative, in addition to high accuracy.

Bayesian neural networks (BayesNNs)~\cite{neal1993bayesian} leverage Bayesian inference to model the epistemic uncertainty, in addition to the default predictive uncertainty, which addresses the limitation of conventional DNNs in  estimating uncertainty.
By representing the weights as probabilistic distributions, BayesNNs provide a principled approach to quantifying their uncertainty, enhancing the robustness and trustworthiness of their predictions in comparison to standard DNNs.
Nevertheless, the benefits of BayesNNs also come with costs:
the high dimensionality of modern BayesNNs introduces prohibitively expensive computation and memory overheads, making the exact Bayesian inference intractable~\cite{gal2016dropout}.

Although various approximation approaches, such as Bayes-by-backprop~\cite{blundell2015weight} and Monte-Carlo Dropout (MCD)~\cite{gal2016dropout}, have been introduced to reduce the algorithmic and hardware complexities of BayesNNs, there are still  two challenges
while deploying BayesNNs in real-world scenarios.
First, BayesNNs generally perform worse than traditional deep ensembles~\cite{lakshminarayanan2017simple} with respect to both accuracy and uncertainty estimation~\cite{ovadia2019can}.
Second, even with the algorithmic approximations, the computational and memory demands of BayesNNs are still much higher than standard DNNs due to Monte-Carlo (MC) sampling, hindering their deployment in demanding applications, especially those with real-time requirements.
While there is extensive research on hardware acceleration for deep learning algorithms, most existing efforts focus on domain-specific hardware~\cite{fowers2018configurable, chen2016eyeriss, fan2022adaptable} or design automation tools~\cite{zhang2018caffeine, fahim2021hls4ml} for standard DNNs such as convolutional DNNs (CNNs)~\cite{krizhevsky2017imagenet, he2016deep} and long short-term memory (LSTM) recurrent DNNs~\cite{hochreiter1997long}.
Hence there are urgent needs for hardware acceleration and algorithmic performance improvements for BayesNNs.



\begin{figure*}[htb]
\centering
\includegraphics[width=0.88\textwidth]{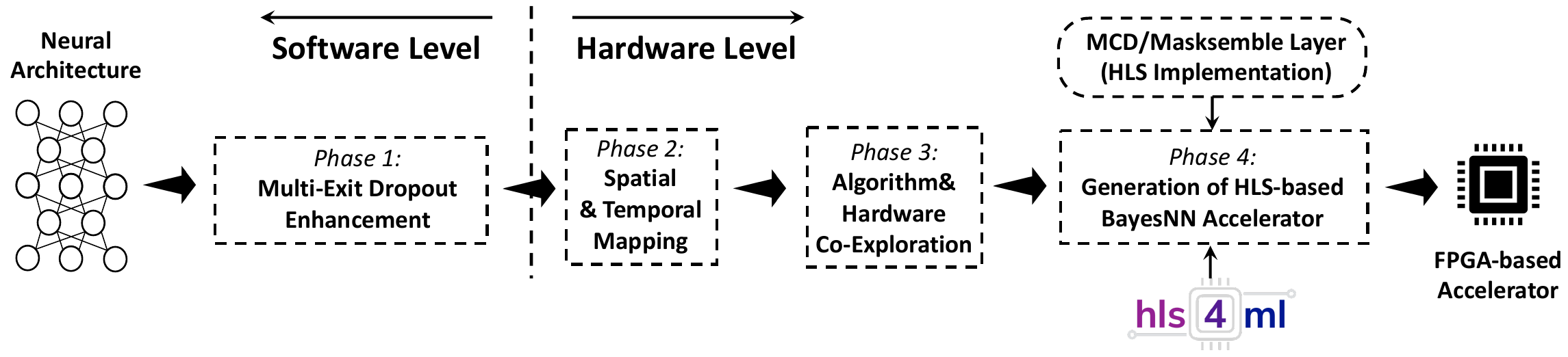}
\vspace{-7mm}
\caption{Overview of our four-phase transformation framework.}
\label{fig:tool_overview}
\end{figure*}

To reduce the algorithmic and hardware barriers of deploying BayesNNs in real-world applications, this paper proposes an algorithm and hardware co-design framework to improve the algorithm and hardware performance of BayesNNs.
At the algorithm level,
we propose a novel multi-exit dropout-based BayesNN that attains low computational and memory overheads while achieving better uncertainty estimation than traditional deep ensembles. 
Furthemore we introduce the hardware support to \textit{Masksemble}~\cite{durasov2021masksembles}-based DNNs and we extend them to multi-exit architectures proposed in this work.
Masksemble is an efficient variant of dropout-based BayesNN without the need for runtime sampling. 
Both approaches fall under the category of dropout-based BayesNNs, each demonstrating unique trade-offs between algorithmic and hardware performance.
At the hardware level, we choose field-programmable gate array (FPGA) technology due to its superior flexibility over application-specific integrated circuit (ASIC) and its potential for achieving higher energy efficiency over graphics processing units (GPUs)~\cite{ma2017optimizing}.
As shown in ~\figref{fig:tool_overview}, we propose a transformation framework to generate high-performance FPGA-based accelerators of multi-exit dropout-based BayesNNs for accurate and efficient uncertainty estimation.
With several novel optimizations such as spatial-temporal mapping and algorithm-hardware co-exploration, the generated accelerators achieve higher energy efficiency than previous hardware implementations.
To facilitate public access to our implementation, we open-source our code at \url{https://github.com/os-hxfan/MCME_FPGA_Acc.git}.

The contributions of this paper can be summarized as follows:
\begin{itemize}[leftmargin=*]

\item Novel multi-exit dropout-based Bayesian neural network (BayesNN) approaches that achieve high quality of uncertainty estimation and high accuracy with low compute and memory overheads (Section \ref{sec:me_dropout}).

\item A transformation framework and optimization strategies including partial dropout, spatial-temporal mapping and algorithm-hardware co-exploration for enhancing algorithm and hardware performance (Section \ref{sec:framework}).

\item A comprehensive evaluation of the proposed approach based
on multiple models and datasets, demonstrating the effectiveness
of our co-design approach (Section \ref{sec:eval}).





\end{itemize}

This work extends our conference publication~\cite{fan2021high}.
The extended material includes:
\textit{1)} multi-exit support on Masksembles to improve their algorithmic performance; \textit{2)} {FPGA-based acceleration of multi-exit Masksemble with optimized implementation to improve hardware performance}; \textit{3)} a more comprehensive evaluation on the quality of uncertainty estimation across multiple models and datasets.

\section{Background and Related Work}\label{sec:background}
\subsection{Bayesian Neural Networks}\label{sec:bnns} In comparison to DNNs, 
BayesNNs demonstrate the capability to effectively \textit{(a)} mitigate overfitting and \textit{(b)} estimate epistemic uncertainty through the utilization of Bayesian inference~\cite{neal1993bayesian}.
In contrast to non-BayesNNs, BayesNNs infer a distribution over their weights through the Bayes rule instead of point-wise weights estimates as encountered in standard DNNs~\cite{neal1993bayesian}.
Despite their advantages, the current BayesNNs~\cite{gal2016dropout} have limited utility in real-world settings because of their high dimensionality which renders the analytical calculation of the aggregated weight distribution computationally infeasible.  

There are two main approaches aiming to approximate the intractable Bayesian inference required by BayesNNs: Markov Chain Monte Carlo (MCMC) and variational inference (VI)~\cite{jospin2022hands}.
MCMC-based methods directly sample from exact posterior distributions, and representative algorithms include Hamiltonian Monte Carlo (HMC)~\cite{neal2011mcmc} and stochastic gradient Langevin Dynamic (SGLD)~\cite{welling2011bayesian} approaches.
Instead of sampling from the exact posterior, VI-based approaches~\cite{gal2016dropout,blundell2015weight} adopt approximate variational distributions with a set of variational parameters.
During training,
the variational parameters are optimized to ensure that their values are as close as possible to the exact posterior weight distribution.


\subsection{Dropout-based Approximations for BayesNNs}\label{sec:bnns}

\subsubsection{MCD-based BayesNNs}\label{sec:bnns_mcd}
Monte-Carlo dropout (MCD)~\cite{gal2016dropout} can be categorized as one of the VI-based approaches that adopt dropout~\cite{srivastava2014dropout} masks to perform efficient Bayesian inference~\cite{jospin2022hands}.
MCD implements a random filter-wise binary mask to remove connections between layers of a DNN. 
The mask values follow a Bernoulli distribution, where the binary random variables take on the value of 0 with a drop rate $p$.
It has been proven that MCD could be interpreted mathematically as approximate Bayesian inference for deep Gaussian processes~\cite{gal2016dropout}.

A key distinction between dropout traditionally employed in standard DNNs~\cite{srivastava2014dropout} and MCD~\cite{gal2016dropout} is that MCD applies dropout during both training and evaluation. 
During evaluation, MCD-based BayesNNs execute multiple forward passes with dropout on and the results are obtained by averaging the output of the multiple MC samples. 
Each forward pass uses an independently generated set of masks, allowing for quantification of the model uncertainty, ultimately enhancing the predictive uncertainty and accuracy. 

\subsubsection{Masksembles}\label{sec:bnns_mask}

By leveraging the predictive power of multiple independent DNNs, deep ensembles~\cite{lakshminarayanan2017simple} can significantly improve accuracy and the quality of uncertainty estimation~\cite{lakshminarayanan2017simple}, while achieving higher robustness against dataset shift~\cite{ovadia2019can}.
However, deep ensembles require the practitioner to train and maintain multiple DNNs in parallel which significantly increases the computational and memory costs during both training and evaluation.

Inspired by MCD-based BayesNNs,
Masksembles~\cite{durasov2021masksembles} train a multi-member deep ensemble inside a single net by using sets of pre-defined dropout masks, effectively reducing the computational and memory overheads in comparison to naive deep ensembles.
Besides, there are another two advantages of Masksembles when compared to MCD-based BayesNNs.
First, since the dropout masks are determined before training and inference, Masksembles eliminate the need for runtime sampling, which effectively reduces the hardware cost.
Second, the overlap and correlation among different dropout masks in Masksembles can be strictly controlled,
allowing it to achieve algorithmic performance similar to that of traditional deep ensembles.




\subsection{Multi-Exit DNNs}\label{sec:multiexit}
Conventional deep learning architectures typically employ a single exit per network to generate predictions.
However, a single-exit architecture exhibits two drawbacks when processing inputs that necessitate only intermediate features extracted from the middle layers.
First, unnecessary computation and memory costs incur as single-exit DNNs always process all the layers until the output layer even when the intermediate features are informative enough for predictions.
Second, certain key features extracted from the intermediate layers might get lost as the network goes deeper, resulting in inaccurate prediction.
To avoid these issues,
multi-exit~\cite{laskaridis2021adaptive} DNNs are introduced that make predictions at various depths of a DNN in a single  forward pass to improve both the algorithm and hardware performance.

While some architectures are specially designed to  support
additional early-exits like \textit{Multi-Scale DenseNet}~\cite{msdnet}, best performance is usually obtained through attaching multiple classifiers to high-performance networks like ResNet~\cite{msdnet}. 
Common choices for where to attach the early exits are after a specific number of floating-point operations (FLOPs) or groupings of convolutional layers \cite{bidistillation, sdn}.
In this paper, we adopt the multi-exit enhancement as an approach to improve the accuracy, uncertainty estimation quality and compute efficiency of BayesNNs.

\subsection{Related Work}

Extensive research has been conducted on DNNs and the use of FPGAs to accelerate them for various applications~\cite{dong2021survey, fan2022optimizing}.
Representative work includes energy-efficient CNN acceleration~\cite{chen2016eyeriss} and FPGA-based real-time AI cloud services~\cite{fowers2018configurable}. 
Significant research also targets design automation for DNNs, like the open source tool {\it hls4ml} supporting an automatic design flow involving high-level synthesis to promote low-power machine learning~\cite{fahim2021hls4ml}.

FPGA-based acceleration of BayesNNs has emerged recently~\cite{wan2021shift}.
Early designs include {\it Bynqnet}, an FPGA-based BayesNNs with quadratic activations for sampling-free uncertainty estimation~\cite{awano2020bynqnet}. 
Efficient FPGA implementations for 2D and 3D convolutional BayesNNs have been proposed~\cite{fan2022fpga}. 
For recurrent Bayesian DNNs,~\cite{ferianc2021optimizing} proposed an FPGA accelerator as well as an algorithmic co-design framework.
Another work is {\it VIBNN}, an FPGA-based accelerator that supports Gaussian distribution-based BaynesNNs sampled at runtime~\cite{cai2018vibnn}. 
Additionally,~\cite{fan2022accelerating} proposed algorithmic and hardware optimizations for BayesNNs, exploiting their structured sparsity and redundant computations.
Lastly,~\cite{ferianc2021effects} explored quantisation in BayesNNs enabling their efficient execution on FPGAs using integer arithmetic.  

In contrast to these approaches, this work extends and differs from the related work in several ways.
First, it proposes a novel multi-exit dropout-based Bayesian DNN, which effectively decreases the computational and memory overhead while achieving high-quality uncertainty estimation and accuracy.
Second, it introduces an automatic tool which translates a software description of the multi-exit BayesNN into a hardware design, executable on an FPGA. 
Third, it introduces several optimization techniques to reduce overall resource consumption and improve the hardware
performance of multi-exit BayesNNs without harming their algorithmic performance. 
These contributions are generalisable to different datasets and DNN architectures, as shown in the experiments, and extensible to previous work mainly through the addition of sampling-based early exits and their hardware consideration. 

\section{Multi-Exit Dropout-based BayesNNs}\label{sec:me_dropout}

\subsection{Multi-Exit Enhancement}

As mentioned in~\secref{sec:bnns}, while both MCD-based BayesNNs and Masksembles demonstrate the potential for efficient predictions and uncertainty estimation, they still suffer from limitations.
On the one hand, MCD-based approximation methods have been criticised due to their inferior performance in uncertainty estimation and confidence calibration when compared to deep ensembles~\cite{ovadia2019can, lakshminarayanan2017simple}.
It has been empirically shown that the introduction of MCD layers after activations in vanilla MCD-based BayesNNs can hamper their predictive power, worsening both their accuracy and uncertainty quantification capabilities~\cite{bayessegnet}.
On the other hand, dropout-based BayesNNs impose a heavy computational burden since obtaining each a prediction necessitates running the entire network multiple times with respect to different dropout masks.
This compute inefficiency hinders their widespread adoption for efficient uncertainty estimation.
To address these drawbacks, this paper proposes a novel multi-exit enhancement for both dropout-based BayesNNs spanning MCD and Masksembles generated masks.
By adopting this approach, we aim to achieve effective and efficient uncertainty estimation, mitigating the limitations of both methods.

\subsubsection{Multi-Exit MCD-based BayesNNs}

\begin{figure}[htb]
\centering
\includegraphics[width=0.49\textwidth]{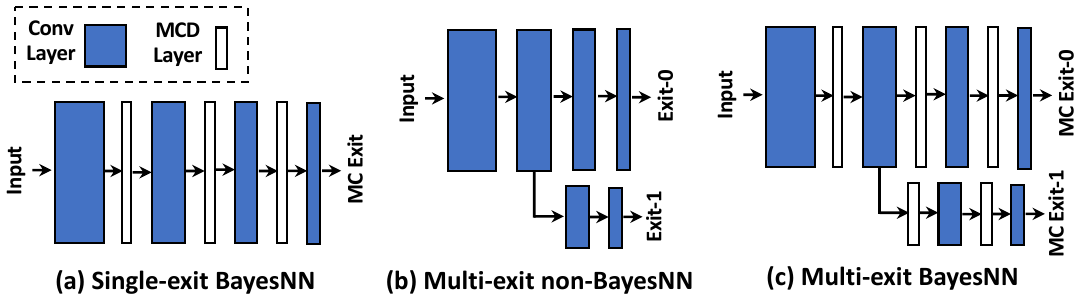}
\vspace{-3.8mm}
\caption{Difference between a single-exit BayesNN, a multi-exit NN and a multi-ext BayesNN.}
\label{fig:multi-exit}
\end{figure}

\figref{fig:multi-exit} presents the network architectures of three distinct approaches: a vanilla MCD-based BayesNN, a multi-exit non-BayesNN, and a multi-exit MCD-based BayesNN proposed in this work.
By adding multiple exits to vanilla MCD-based BayesNNs, we propose multi-exit MCD-based BayesNNs, as depicted in~\figref{fig:multi-exit}(c).
In contrast to the traditional single-exit MCD-based BayesNNs, our multi-exit MCD-based approach makes predictions from exits at different depths of the network, which
effectively improves the quality of uncertainty estimation as well as hardware efficiency, as demonstrated in Section~\ref{subsec:eff_memc}.
Furthermore, when compared to multi-exit non-BayesNNs, our proposed approach has the advantage of generating arbitrary prediction MC samples with the use of MCD layers,
improving the flexibility for uncertainty estimation.
An intriguing aspect of multi-exit MCD-based BayesNNs lies in the capability of capturing the uncertainty across different network depths.
This stems from the utilization of diverse intermediate features extracted from different stages of the network to enable the network to make diverse predictions.

\subsubsection{Multi-Exit Masksembles}
Although the use of MCD layers enables flexibility in making arbitrarily many predictions, it also introduces hardware overhead due to the frequent Bernoulli sampling to generate the masks.
To provide a hardware-efficient alternative, we propose multi-exit Masksembles to replace MCD layers with Masksemble layers.
To avoid the highly correlated predictive results across multiple exits, we adopt the mask scale parameter~\cite{durasov2021masksembles} to control the overlap among different pre-defined masks.
There are two distinct computational differences when comparing MCD-based and Masksemble-based approaches.
First, by adopting pre-defined binary masks, multi-exit Masksembles eliminate the need for sampling during runtime.
As the locations of zeros are fixed, it provides us with the opportunity for designing efficient hardware accelerators to intelligently skip redundant computation associated with zero values, as discussed in~\secref{sebsec:gen_acc}.
Second, MCD-based method applies dropout in the channel granularity, while Masksemble layer adopts point-wise masks with more fine-grained dropout granularity.
These two difference lead to distinct hardware design requirements while accelerating multi-exit MCD-based BayesNNs and multi-exit Masksembles.

This paper treats both MCD layers and Masksembles layers as two distinct dropout layers, each exhibiting specific trade-offs among accuracy, uncertainty and hardware performance.
To fulfil the diverse needs of different users,we propose a co-design framework dedicated to optimizing  the dropout layers, as elaborated in~\secref{sec:framework}.
This optimization enables users to tailor the final network for their target applications, ultimately leading to efficient prrediction and uncertainty estimation for various scenarios.





\subsection{Partial Dropout}\label{subsec:partial_dropout}

Applying dropout after every convolution incurs large computational overhead since it requires running the whole network multiple times to get the predictions.
Inspired by~\cite{kristiadi2020being, kendall2015bayesian, fan2021high}, we propose partial dropout for both multi-exit Masksembles and multi-exit MCD-based BayesNNs.
Rather than applying dropout to every learnable layer~\cite{gal2016dropout}, we insert dropout layers starting from exits towards the input part of the network.
We refer to the layers without dropout applied as the non-Bayesian component of the network.
By placing dropout layers closer to each exit, fewer computations are required since the non-Bayesian results can be cached and reused for different prediction samples. 

With partial dropout applied, both multi-exit MCD-based BayesNN and multi-exit Masksembles can be interpreted as ensembles of approximated BayesNNs built upon the non-Bayesian component feature extractor.
Given an $M$-exit architecture with inputs $\mathbf{X}$, our approach first maps the data from input space into feature space by using $f_{i}(X)$, where $f_{i}(.)$ denotes the feature extractor of each exit with $1 \leq i \leq M$.
Built upon the features extracted by $f_{i}(X)$,
each exit then adopts the dropout-based Bayesian approach through either MCD or Masksembles layer to make predictions.
The final result ensembles predictions from different approximated BayesNNs with multiple exits.





\subsection{Compute Efficiency}


We demonstrate that our proposed multi-exit dropout-based BayesNNs have higher compute efficiency over single-exit BayesNNs in making predictions.
Given that the FLOPs of the non-Bayesian feature extractor and all the exits are $FLOP_{main}$ and $FLOP_{exit}$ respectively.
To get a single MC sample, it is necessary to run the entire BayesNN end-to-end and the computational cost of running $N_{sample}$ MC samples can be formulated as:
\begin{equation}\label{eq:cost_single_exit}
    N_{sample} \times (FLOP_{main} + FLOP_{exit}).
\end{equation}
In contrast, the required FLOPs of an $N_{exit}$ multi-exit dropout-based BayesNN to get the same number of predictions is:
\begin{equation}\label{eq:cost_multi_exit}
    FLOP_{main}+ \frac{N_{sample}}{N_{exit}} \times FLOP_{exit}.
\end{equation}
The reduction rate is given by dividing Equation~\ref{eq:cost_single_exit} by Equation~\ref{eq:cost_multi_exit},
\begin{equation}
    \frac{1 + \alpha}{\frac{1}{N_{sample}} +  \frac{\alpha}{N_{exit}}},
\end{equation}
where $\alpha = \frac{FLOP_{exit}}{FLOP_{main}}$.
The reduction rate varies by different multi-exit architectures, depending on $N_{sample}$, $N_{exit}$ and $\alpha$.

Section~\ref{sec:multiexit} discusses the wide variety of possible methods in which multi-exit networks can be created and trained. 
In this work, the exit branches are placed according to the approach used in~\cite{bidistillation}.
Semantic groupings are formed for each network, splitting the network architecture into ``blocks" separated by pooling layers. 
An exit branch is then placed after each of these blocks. 
In order to allow for more direct validation of the work performed in this paper, the bidirectional distillation training method~\cite{bidistillation} is used.

\section{Transformation Framework}\label{sec:framework}
\subsection{Framework Overview}

This section describes the proposed transformational framework presented in~\figref{fig:tool_overview}.
It comprises multiple steps: \textit{(1)} adaptation of the architecture and evaluation protocol for multi-exit dropout, \textit{(2)} spatial and temporal mapping optimization, \textit{(3)} algorithm and hardware co-exploration and \textit{(4)} generation of FPGA-based accelerators for BayesNNs using High-Level Synthesis (HLS). 

Given a neural architecture description as an input, the first phase applies early-exits enhanced either with MCD~\cite{gal2016dropout} or Masksembles~\cite{durasov2021masksembles} approaches, and decides the number of MC samples according to the user-specified requirements.
The second phase exploits spatial and temporal processing in BayesNNs and implements optimisations to improve the runtime hardware performance.
The third phase involves algorithm and hardware co-exploration to optimize design parameters such as bitwidth and execution strategies depending on both the network architecture as well as the available hardware resources in terms of DSPs or memory budget. 
Given the network architecture as well as the obtained hardware parameters, the last phase produces the final HLS-based hardware implementation executable on an FPGA. 
We adopt the design flow and HLS template of common NN layers from \textit{hls4ml}~\cite{fahim2021hls4ml} and we develop an HLS-based implementation of MCD/Masksembles layers and \textit{Keras}-to-HLS conversion into the design flow in order to generate the executable hardware implementation.

\subsection{Multi-Exit Dropout: Phase 1}

\begin{figure}[htb]
\centering
\includegraphics[width=0.4\textwidth]{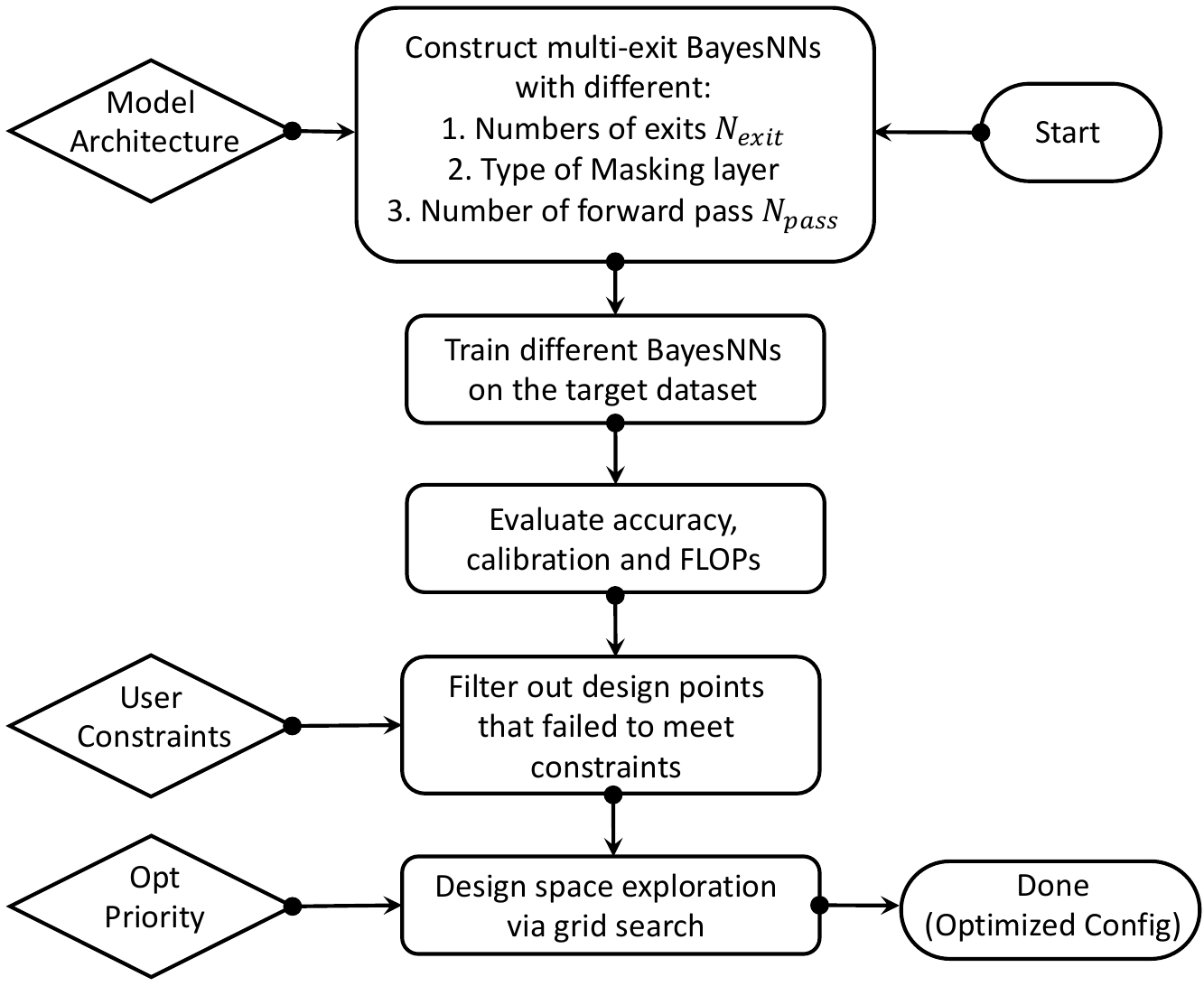}
\caption{Optimization flow.}
\label{fig:flow_chart}
\end{figure}

Multi-exit dropout phase optimizes the design parameters for multi-exit dropout-based BayesNNs, including: the number of exits $N_{exit}$, the number of forward passes $N_{pass}$, the type of dropout layers and the associated dropout parameters, and the total number of MC samples $N_{sample}$.
The parameters trade-off software and hardware performance, namely accuracy, calibration and latency.
For instance, the total number of MC samples $N_{sample}$ from a multi-exit dropout BayesNN with $N_{exit}$ exits and $N_{pass}$ passes is calculated as $N_{sample} = N_{pass} \times N_{exit}$.
Higher values of $N_{exit}$ and $N_{pass}$ can improve accuracy and calibration but also increase the total $N_{sample}$ count.
This leads to worse hardware performance because more forward passes through the network or the exits are needed, increasing computational and memory demands.
To optimize these hyperparameters for different applications and architectures, balancing both algorithm and hardware metrics, we propose a multi-exit dropout optimization flow as shown in~\figref{fig:flow_chart}.

The multi-exit dropout optimization starts by constructing different dropout-based BayesNNs based on the default input architecture provided by the user.
By inserting $N_{exit}$ exits with either MCD or Masksembles layers, different BayesNNs candidates are constructed and trained on the target dataset.
After training, we evaluate each model with respect to software and hardware metrics like accuracy, calibration, and FLOPs.
Models that do not meet specified constraints on these metrics, given by the users, are filtered out.
Then, according to the optimization metric priority, design space exploration is performed to find the optimal design configuration via grid search.
The priority can be set with respect to a single or multiple metrics, specified by the user e.g. accuracy, calibration and the number of FLOPs.
The final optimized design is fed into the next stage for hardware design generation.

\subsection{Spatial and Temporal Mappings: Phase 2}\label{subsec:spt_temp_map}

\begin{figure}[htb]
\centering
\vspace{-2.0mm}
\includegraphics[width=0.49\textwidth]{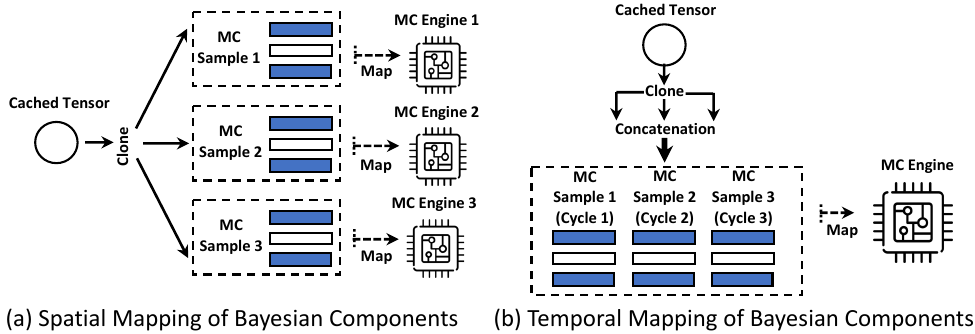}
\vspace{-5.0mm}
\caption{Spatial and temporal mappings for Bayesian components.}
\vspace{-1.0mm}
\label{fig:spt_temp_map}
\end{figure}

Bayesian components with either MCD or Masksemble layers require multiple forward passes to generate MC samples from the predictive distribution.
Compared with conventional non-Bayesian NNs, the Bayesian components exhibit concurrency along the MC sampling dimension.
This creates new opportunities for parallelism compared to non-Bayesian networks.
Therefore, we propose two mapping hardware optimisation strategies, spatial and temporal, to accelerate Bayesian NNs, which are illustrated in~\figref{fig:spt_temp_map}.

In both mapping strategies, the data generated from the last non-Bayesian layer are cached and cloned.
As shown in~\figref{fig:spt_temp_map}(a), spatial mapping uses separate MC engines for each sample.
Although spatial mapping effectively reduces latency by enabling parallel sampling, it also significantly increases computational resource usage when the number of MC samples becomes high.
To alleviate this issue, we propose temporal mapping that shares one MC Engine among multiple MC samples.
As shown in~\figref{fig:spt_temp_map}(b), the cloned copies of the cached data are concatenated before feeding it into the shared engine.
The engine then maps the computation of different MC samples one by one onto a single MC Engine. 
Our approach optimizes the mix of spatial and temporal mappings to meet different latency and resource constraints.

\subsection{Algorithm and Hardware Co-Exploration: Phase 3}
Our hardware accelerator has various design parameters, such as the implementation strategy used in \textit{hls4ml}, layer reuse factors and Bayesian mapping approaches.
On the algorithm side, given an input model architecture, we can optimize hyperparameters like the number of channels for different layers and bitwidths for activations/weights.
We co-explore both algorithm and hardware parameters using grid search to optimize the design with similar algorithm accuracy to defaults.
To reduce search costs, we experiment with heuristics such that the bitwidth for activations or weights is chosen from $\{4, 6, 8, 16\}$ bits and the channels selected from $\{C, \frac{C}{2}, \frac{C}{4}, \frac{C}{8}\}$ with $C$ being the original number of channels. 
Users can also define other dimensions for the search space. 
This joint optimization allows customizing algorithmic and hardware configurations for different constraints.

\subsection{Generation of FPGA-based Accelerator: Phase 4}\label{sebsec:gen_acc}

We generate HLS-based accelerators using \textit{hls4ml} design and our custom templates for MCD/Masksembles layers.
The accelerators are synthesized and implemented in Vivado HLS to produce FPGA bitstreams for deployment.
The pseudocode of HLS-based implementation of MCD and Masksemble layers are presented in Algorithm~\ref{algrm:hls_mcd} and~\ref{algrm:hls_mask}, respectively.
We apply optimizations like pipelining and caching, as described in the previous Section, to improve performance. 
In both implementations, the HLS directive \textit{HLS PIPELINE} is used to improve the overall performance through pipelining.
We cache the temporary result in the variable \textit{temp}, before generating the final outputs.
The hardware receives layer inputs and streams outputs to the next layer.
For MCD, the dropout rate $P_{dropout}$ is a specified parameter by the user at the beginning of running each model.
A multiplexer selects between zero or the input scaled by the rate based on comparing to a random number.
The control signal of the multiplexer is generated by comparing $P_{dropout}$ with \textit{uniform\_random}.
To support the MCD layer with arbitrary $P_{dropout}$, a random number generator is used in our design to generate \textit{uniform\_random}.
For Masksembles, the masks are provided as inputs, avoiding sampling in hardware. 
The inputs with mask values being one are passed through to the outputs.

\begin{algorithm}
\caption{Pseudocode of MCD layer}\label{algrm:hls_mcd}
  \begin{algorithmic}[1]
   \State \textbf{Input}: input[$dropout\_size$], keep\_rate
   \State \textbf{Output}: output[$dropout\_size$]
    \For{($i$ from $0$ to $dropout\_size$)} \Comment{\#pragma PIPELINE}
        \State temp = input[i]
        \State uniform\_random = random\_number\_generator()
        \If{(uniform\_random $>$ keep\_rate)} temp = 0
       \EndIf
       \State output[i] = temp * keep\_rate
    \EndFor
  \end{algorithmic}
\end{algorithm}

\begin{algorithm}
\caption{Pseudocode of Masksembles layer}\label{algrm:hls_mask}
  \begin{algorithmic}[1]
   \State \textbf{Input}: input[$mask\_size$], mask\_index, \\ \qquad \quad generated\_masks[$mask\_num$][$mask\_size$]
   \State \textbf{Output}: output[$mask\_size$]
    \For{($i$ from $0$ to $mask\_size$)} \Comment{\#pragma PIPELINE}
        \State mask\_value = generated\_masks[$mask\_index$][$i$]
        \If{(mask\_value $==$ 0)}
        \State output[i] = 0
        \Else
        \State output[i] = input[i]
        \EndIf
    \EndFor
  \end{algorithmic}
\end{algorithm}


\section{Experiments and Evaluation}\label{sec:eval}
Our optimization framework is implemented in Python $3.8.12$, PyTorch $1.11.0$, and Keras $2.9.0$.
We use Vivado-HLS $2020.1$ for hardware implementation.
QKeras $0.9.0$ is used for quantization.
The latency and resource consumption are obtained from C-synthesis reports provided by Vivado-HLS.
Vivado $2020.1$ is used to run place and route for the final designs.
We set Xilinx Kintex XCKU115 as our target FPGA board.
All the designs are optimized by our spatial-temporal mapping and algorithm-hardware co-exploration to ensure they can be fitted into the target platform.

\subsection{Resource Cost of Being Bayesian}

\begin{figure*}[tbh!]
\centering
\includegraphics[width=0.99\textwidth]{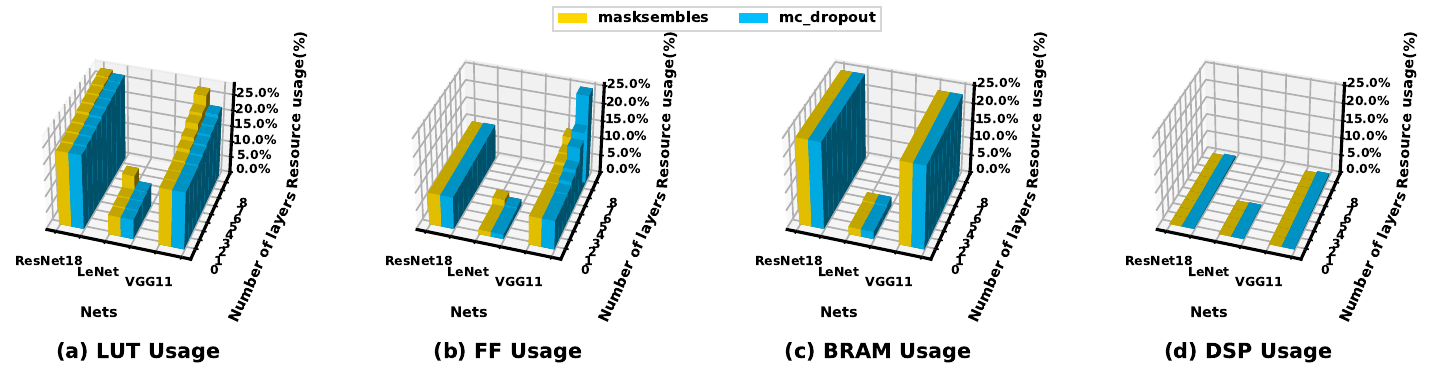}
\caption{{Resource consumption of mask-based BayesNNs with LeNet, ResNet18 and VGG11 as network backbones. The quantization and custom number of channels are applied to fit models onto FPGAs.}}
\label{fig:resource_cost}
\end{figure*}

Inserting dropout layers transforms
conventional DNNs into BayesNNs, enabling reliable uncertainty estimation required by various safety-critical applications.
To quantitatively investigate the hardware overhead imposed by the transformation,
we evaluate the resource consumption of Bayesian accelerators against their non-Bayesian counterparts.
Three BayesNNs and datasets are used in our experiments, i.e., \textit{LetNet5} on MNIST, \textit{ResNet-18} on CIFAR-10, and \textit{VGG-11} on SVHN.
As we aim to evaluate the resource cost of being Bayesian, all the models use single-exit to eliminate the hardware overhead introduced by the multi-exit optimization.
We generate different Bayesian accelerators with distinct numbers of dropout layers using our proposed design flow from~\secref{sec:framework}.
For non-Bayesian accelerators, we set the number of dropout layers as zero.
In order to fit BayesNNs onto FPGA,
we apply quantization and custom channel numbers to ease the memory requirements.
To further reduce compute resource consumption,
we adopt temporal mapping on all the hardware designs.

\begin{figure}[bh!]
\centering
\includegraphics[width=0.49\textwidth]{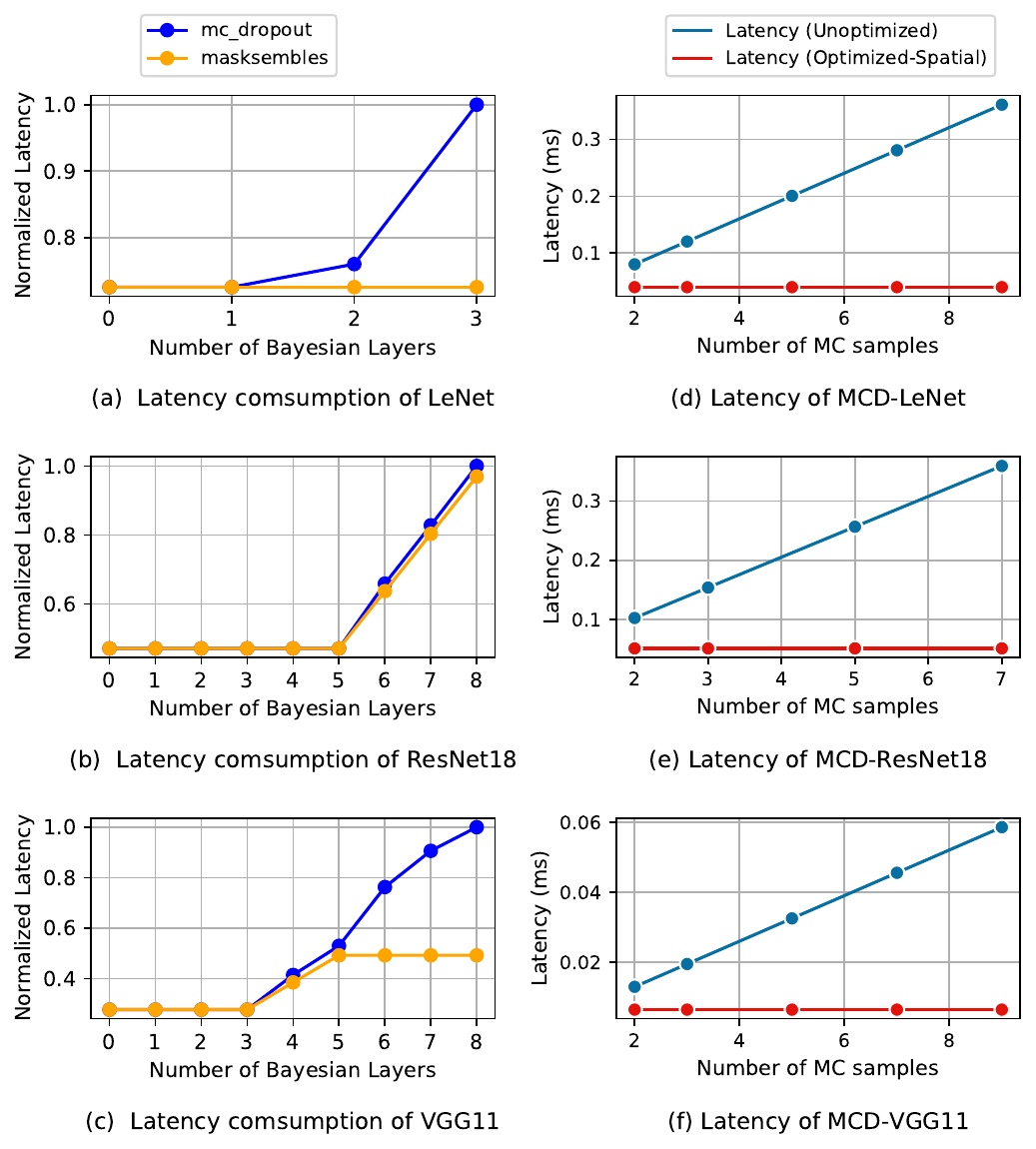}
\caption{{Latency reduction of different hardware optimization techniques.}}
\label{fig:lat_cost}
\end{figure}

\figref{fig:lat_cost} shows the resource consumption of Block RAM (BRAM), DSP, Flip-Flop (FF) and Look-up Tables (LUTs) 
We implement two different dropout types, MCD and Masksembles with varied numbers of dropout layers for each model.
As can be observed in all three models,
the BRAM and DSP usage stays almost the same across different numbers of dropout layers and dropout types.
The reason is that dropout layers do not contain compute and memory-intensive operations.
The designs of both MCD and Masksemble layers can be implemented by mainly just using logic resources.
In contrast, an increasing trend can be observed in both FF and LUT consumption when more dropout layers are inserted.
The most significant increase is found on MCD-based Bayes-VGG, where nearly $13$\% more FF resources are utilized for the insertion of $8$ dropout layers. 
However, this overhead is caused by inserting MCD layers after every convolution.
With our proposed partial dropout in~\secref{subsec:partial_dropout}, the LUT and FF resource overheads of one MCD layer are just around $1\% \thicksim 2\%$, demonstrating the resource-efficiency of our co-design approach.

\begin{table*}[tb]
\centering
\caption{Performance comparison among SE CNNs, MCD-ME, and Mask-ME with 32-bit floating point (FP32).}
\vspace{-2.5mm}
\vspace{0.3cm}
\label{tb:multi-exit}
\setlength\tabcolsep{1pt}
\scalebox{0.9}{
\begin{tabular}{C{2.cm}|C{2.2cm}|C{2.5cm}C{2.5cm}C{2.5cm}C{2.5cm}C{2.5cm}C{2.5cm}}
\toprule
\multirow{2}{*}{\textbf{Network}}&\multirow{2}{*}{\textbf{Approach}} & \multicolumn{2}{c|}{Acc-Opt} & \multicolumn{2}{c|}{ECE-Opt} & \multicolumn{2}{c}{aPE-Opt} \\
\cmidrule{ 3 - 8 }
& & Accuracy & FLOPs & ECE & FLOPs & aPE & FLOPs \\
\midrule
\multirow{3}{*}{\textbf{\textit{Bayes-ResNet}}}& SE & $0.752 \pm 0.002$ & $\boldsymbol{1.00}$ & $0.0840 \pm 0.0008$ & $1.00$ & $2.048 \pm 0.643$ & $1.00$ \\
\cmidrule{ 2 - 8 }
& MCD+ME & {$\boldsymbol{0.776 \pm 0.001}$} & $1.019 \pm 0.004$ & $\boldsymbol{0.014 \pm 0.001}$ & ${0.672 \pm 0.003}$ & $\boldsymbol{2.367 \pm 0.847}$ & $0.586$ \\
\cmidrule{ 2 - 8 }
& Mask+ME & {${0.764 \pm 0.004}$} & $1.032 \pm 0.006$ & ${0.016 \pm 0.001}$ & $\boldsymbol{0.605 \pm 0.001}$ & ${2.116 \pm 1.301}$ & $\boldsymbol{0.462}$ \\
\midrule
\multirow{3}{*}{\textbf{\textit{Bayes-VGG}}} & SE & $0.693 \pm 0.002$ & $\boldsymbol{1.00}$ & $0.165 \pm 0.006$ & $1.00$ & $1.287 \pm 0.578$ & $1.00$ \\
\cmidrule{ 2 - 8 }
& MCD+ME & $\boldsymbol{0.747 \pm 0.001}$ & $0.982$ & $\boldsymbol{0.017 \pm 0.001}$ & $\boldsymbol{0.45 \pm 0.02}$ & ${2.664 \pm 0.721}$ & $\boldsymbol{0.343}$ \\
\cmidrule{ 2 - 8 }
& Mask+ME & ${0.741 \pm 0.001}$ & $0.982$ & ${0.019 \pm 0.003}$ & $0.49 \pm 0.05$ & $\boldsymbol{2.741 \pm 0.867}$ & $0.419$ \\
\bottomrule
\end{tabular}
}
\end{table*}

\begin{table*}[t]
\centering
\caption{Performance comparison of our final FPGA designs with CPU, GPU, and other FPGA-based implementations. \vspace{0.1cm}}
\label{tb:compare_sota}
\setlength\tabcolsep{1pt}
\scalebox{0.88}{
\begin{tabular}{C{3.99cm}C{2.5cm}C{2.5cm}C{2.5cm}C{2.0cm}C{2.0cm}C{2.0cm}C{2.0cm}}
\toprule
{}&{\bf CPU}&{\bf GPU}&{\bf ASPLOS'18~\cite{cai2018vibnn}} &{\bf \bf DATE'20~\cite{awano2020bynqnet}}& {\bf DAC'21~\cite{fan2021high}}&{\bf TPDS'22~\cite{fan2022accelerating}}& {\bf Our Work}\\
\midrule 
{\bf Platform}& {Intel Core i9-9900K} &{NVIDIA RTX 2080 } &{Altera Cyclone V} & {Zynq XC7Z020}&{Arria 10 GX1150}&{Arria 10 GX1150}  &{XCKU 115} \\
\midrule
{{\bf Frequency} (MHz)} & 3600& 1545& 213 & 200 &{225}& {220}&{\SI{181}{}}\\\midrule
{{\bf Technology}} & {14 nm} & {12 nm} & {28 nm}  & {28 nm} &{20 nm}& {20 nm}&{20 nm}\\\midrule
{{\bf Power} (W)} & {205}& {236}& {6.11} & {2.76} &{45.00}& {43.6}&{4.383}\\\midrule
{\bf Latency} (ms)& {1.26}& {0.57}& {5.5}& {4.5}& {0.42}& {0.32}&{0.89}\\\midrule
{\bf Energy Efficiency} (J/Image) & {0.258}& {0.134}& {0.033}& {0.012}& {0.019}& {0.014}&{0.004}\\\bottomrule
\end{tabular}}
\vspace{0.0mm}
\end{table*}

\subsection{Latency Reduction of Masksembles and Spatial Mapping}
By adopting a set of pre-defined dropout masks,
Masksembles eliminate the need for runtime Bernoulli sampling, exhibiting higher hardware efficiency than MCD-based BayesNNs.
To quantitatively evaluate the hardware performance improvement of Masksembles compared with MCD-based BayesNNs,
we generate different accelerators for both approaches with distinct numbers of dropout layers.
We set the \textit{hls4ml} optimization strategy as "\textit{Resource}" to ensure the generated accelerators can be fitted onto the target FPGA board.
\figref{fig:lat_cost}(a)$\sim$(c) show the normilzed latency of Bayes-LetNet5, Bayes-ResNet and Bayes-VGG-11, respectively.
As it can be observed,
the use of Masksembles layers can effectively reduce the latency of the generated accelerators.
This latency reduction is more significant on Bayes-LetNet and Bayes-VGG with a larger number of dropout layers.

Spatial mapping is another optimization that we propose to reduce latency when more hardware resources are available on the FPGA.
To demonstrate the effectiveness of spatial mapping in reducing latency,
we evaluate accelerators with and without spatial mapping optimization.
As the type of dropout layer will not affect this demonstration,
we take MCD-based BayesNNs as examples and apply partial masking by only inserting MCD after the last convolutional layer.
The \textit{hls4ml} optimization strategy is set as "\textit{Latency}" to ensure best latency performance.
\figref{fig:lat_cost}(d)$\sim$(e) show the latency results of both optimized and un-optimized accelerators with different numbers of MC samples on three network backbones.
As can be seen, the latency of an unoptimized accelerator increases linearly with the increase of MC samples.
In contrast,
the latency of spatial-optimized accelerators stays almost the same when the number of MC samples increases, demonstrating the effectiveness of spatial mapping.
The improvement of spatial mapping stems from its mechanism of deploying multiple physical cores to compute MC samples in parallel.

\subsection{Effect of Multi-Exit Enhencement}\label{subsec:eff_memc}
To demonstrate the advantage of multi-exit BayesNNs over the baseline approaches,
we evaluate two multi-exit models, \textit{VGG19} and \textit{ResNet18} for image classification.
Cifar100 dataset, a curated subset of a larger dataset scraped from the web containing photo-realistic tiny $32 \times 32$ images with a single main object, is used in this experiment.
We use Expected Calibration Error (ECE)~\cite{ovadia2019can} as a metric to evaluate calibration ability.
To measure predictive uncertainty, we measure the average predictive entropy (aPE) across a Gaussian noise dataset with the same mean and variance as the training data~\cite{fan2021high}.

We compare three different implementations: \textit{i)} Single-exit model with only one exit at the end of the network (SE). There is no MCD or Multi-Exit applied, which is the original implementation of both the  \textit{ResNet-18} and \textit{VGG-19}. 
\textit{ii)} MCD-based BayesNN with multi-exit (MCD + ME). The MCD is applied to every exit of the network.
\textit{iii)} Maskembles-based BayesNN with multi-exit (Mask + ME). The Masksembles layer is applied to every exit of the network.
Stochastic gradient descent (SGD) is used with a weight decay of $5 \times 10^{-4}$, an initial learning rate of $0.1$ and a momentum of $0.9$, along with a batch size of $64$. 

As discussed previously, the usage of too many dropout layers in a BayesNN can overregularize the network and adversely affect performance. 
However, there is no standard method to find the best balance
between the level of dropout and calibration.
Therefore, a small grid search is performed over different dropout configurations.
For MCD layers, we optimize the dropout rates from 0.125, 0.25, 0.375 and 0.5.
The scale parameter of the Masksemble layer is selected from 3, 4, 5 and 6.
Similarly, the threshold of confidence-based exiting~\cite{sdn} which optimally balances the computational cost and the network performance is found through testing the same thresholds as in \cite{sdn}: 0.1, 0.15, 0.25, 0.5, 0.6, 0.7, 0.8, 0.9, 0.95, 0.99, 0.999. It is noted that two sets of results from performing confidence-based exiting are calculated, using the predictions at each exit or the largest possible ensemble at each exit respectively. Each ensemble is an equally weighted average of the predictions from each exit, as in \cite{ee}.

The grid search covers all combinations of the above dropout configurations, which is applied to the networks. The predictions from each of the exits and the ensembles formed by averaging the results from each exit are calculated, alongside the predictions from confidence exiting. 
The results are presented in~\tabref{tb:multi-exit}.
As the dropout rate of MCD and the confidence threshold of multi-exit may affect accuracy, calibration and uncertainty,
three configurations for each implementation and model are reported: those that achieve the highest accuracy (\textit{Acc-Opt}), the lowest ECE (\textit{ECE-Opt}), and the highest aPE (\textit{aPE-Opt}).
For each configuration, we also calculate the FLOPs as a fraction of the SE implementation.

On \textit{ResNet18},
our approach, MCD + ME, improves the accuracy by $2.4$\% $\pm 0.002$\% with only $0.019$ times more FLOPs compared with the SE implementation.
In \textit{ECE-Opt} and \textit{aPE-Opt},
both MCD + ME and Mask + ME achieve lower ECE and higher aPE than SE approach.
A similar trend can also be observed in \textit{VGG-19}. 
These results show that multi-exit dropout-based BayesNNs can lead to better calibration and uncertainty estimation while costing or fewer FLOPs.

\subsection{Comparison with CPU, GPU, and FPGA implementations}
To demonstrate the energy efficiency of our approach,
we also compare it against CPU, GPU, and other FPGA-based implementations.
The comparison uses MNIST dataset since it is the most common dataset across different work~\cite{cai2018vibnn, awano2020bynqnet, fan2021high, fan2022accelerating}.
As both~\cite{cai2018vibnn} and~\cite{awano2020bynqnet} do not support \textit{Bayes-LeNet5},
we use their reported throughput (GOP/s) to estimate their performance on \textit{Bayes-LeNet5}.
The performance is obtained by using three MC samples.
Both CPU and GPU performance are quoted from the vanilla implementations of MCD-based BayesNNs in~\cite{fan2022accelerating}.
Although there are some other BayesNN accelerators~\cite{wan2020fast,wan2021shift},
they do not report any end-to-end latency and energy consumption.

As shown in~\tabref{tb:compare_sota},
our design achieves $65$ and  $33$ times higher energy efficiency than CPU and GPU implementations, despite the FPGA adopting 20nm technology while the CPU adopting 14nm technology and the GPU adopting 12nm technology.
Our accelerator also shows lower latency and better energy efficiency than both~\cite{cai2018vibnn} and~\cite{awano2020bynqnet}.
Although both~\cite{fan2021high} and~\cite{fan2022accelerating} are faster than our design,
they consume much higher energy due to the high resource utilization and frequent data transfer between on-chip and off-chip memory, leading to nearly $5$ and $4$ times lower energy efficiency than our design.
Also, compared with their Verilog-based implementations, our HLS-based accelerator has advantages in development time~\cite{pelcat2016design}, which can improve designer productivity and can facilitate extending our approach to cover other NNs such as LSTM~\cite{hochreiter1997long}.
\tabref{tb:power_breakdown} provides the power consumption breakdown obtained from the Xilinx Power Estimator (XPE) tool after place and route.
For the MCD-based BayesNNs, the dynamic power occupies $70\%$ of the total power.
The logic\&signal and IO consume most of the dynamic power,
accounting for $31\%$ and  $17\%$, respectively. This pattern is also obseverd in the mask-based BayesNNs.
The high IO power consumption results from our spatial mapping strategy with multiple MC engines running in parallel.

\begin{table}[t]
\centering
\vspace{-1.0mm}
\caption{Power breakdown of our FPGA-based accelerator.}
\label{tb:power_breakdown}
\setlength\tabcolsep{1pt} 
\scalebox{0.9}{
\begin{tabular}{C{0.9cm}|C{1.4cm}|C{1.1cm}|C{1.0cm}|C{1.0cm}|C{0.7cm}|C{0.8cm}|C{1.0cm}|C{1.0cm}}
\toprule
\multicolumn{2}{c|}{}&\multicolumn{5}{c|}{\textbf{Dynamic} (W)} & \multirow{3}{*}{\textbf{Static}} & \multirow{3}{*}{\textbf{Total}} \\ \cmidrule{3-7}
\multicolumn{2}{c|}{}&\multirow{2}{*}{Clocking} &  {Logic\&} & \multirow{2}{*}{BRAM} &  \multirow{2}{*}{IO} &\multirow{2}{*}{DSP} & \\ 
\multicolumn{2}{c|}{}& & {Signal} & & & &  \\ \midrule
{\textit{\textbf{Bayes}}}&\textbf{Used} & 0.365  & 1.407  & 0.421 & 0.728 & 0.166 & 1.295 & 4.383 \\ \cmidrule{2-9}
{\textit{\textbf{MCD}}}&\textbf{Percentage} & 8\% & 31\% & 10\%  & 17\% & 4\%& 30\%& 100\%\\
\cmidrule{1-9}
{\textit{\textbf{Bayes}}}&\textbf{Used} & 0.355  & 1.235  & 0.514 & 0.685 & 0.153 & 1.294 & 4.235 \\ \cmidrule{2-9}
{\textit{\textbf{Mask}}}&\textbf{Percentage} & 8\% & 29\% & 12\%  & 17\% & 3\%& 31\%& 100\%\\
\bottomrule
\end{tabular}}
\vspace{-1mm}
\end{table}

\section{Conclusion}

This paper proposes an algorithm and hardware co-design approach for accelerating dropout-based multi-exit Bayesian Neural Networks (BayesNNs).
On the algorithm side,
we propose novel multi-exit dropout-based BaeysNNs that achieve high algorithmic performance with low computational and memory overhead.
At the hardware level, we introduce a transformation framework to generate FPGA-based accelerators for multi-exit dropout-based BayesNNs.
Multiple optimizations including the mix of spatial and temporal mappings are proposed to further improve the overall performance of dropout-based BayesNNs.
Comprehensive experiments demonstrate that our approach achieves higher algorithmic and energy efficiency than state-of-the-art designs.
To facilitate public access to BayesNNs hardware
accelerators,
we have made our code accessible as an open-source resource at: \url{https://github.com/os-hxfan/MCME_FPGA_Acc.git}
In the future, we aim to automate the transformation framework, extend support for attention-based BayesNNs, and incorporate capabilities such as run-time reconfiguration.


\section*{Acknowledgement}{
The support of UK EPSRC grants (UK EPSRC grants EP/L016796/1, EP/N031768/1, EP/P010040/1, EP/V028251/1 and EP/S030069/1) is gratefully acknowledged.
}

\bibliographystyle{IEEEtran}
\bibliography{refs}

\begin{thebibliography}{10}
\providecommand{\url}[1]{#1}
\csname url@samestyle\endcsname
\providecommand{\newblock}{\relax}
\providecommand{\bibinfo}[2]{#2}
\providecommand{\BIBentrySTDinterwordspacing}{\spaceskip=0pt\relax}
\providecommand{\BIBentryALTinterwordstretchfactor}{4}
\providecommand{\BIBentryALTinterwordspacing}{\spaceskip=\fontdimen2\font plus
\BIBentryALTinterwordstretchfactor\fontdimen3\font minus \fontdimen4\font\relax}
\providecommand{\BIBforeignlanguage}[2]{{%
\expandafter\ifx\csname l@#1\endcsname\relax
\typeout{** WARNING: IEEEtran.bst: No hyphenation pattern has been}%
\typeout{** loaded for the language `#1'. Using the pattern for}%
\typeout{** the default language instead.}%
\else
\language=\csname l@#1\endcsname
\fi
#2}}
\providecommand{\BIBdecl}{\relax}
\BIBdecl

\bibitem{dong2021survey}
S.~Dong \emph{et~al.}, ``A survey on deep learning and its applications,'' \emph{Computer Science Review}, vol.~40, p. 100379, 2021.

\bibitem{otter2020survey}
D.~W. Otter \emph{et~al.}, ``A survey of the usages of deep learning for natural language processing,'' \emph{IEEE Transactions on Neural Networks and Learning Systems}, vol.~32, no.~2, pp. 604--624, 2020.

\bibitem{fan2021high}
H.~Fan \emph{et~al.}, ``High-performance {FPGA}-based accelerator for {B}ayesian neural networks,'' in \emph{Proceedings of the 2021 ACM/IEEE Design Automation Conference (DAC)}.\hskip 1em plus 0.5em minus 0.4em\relax IEEE, 2021, pp. 1--6.

\bibitem{leibig2017leveraging}
C.~Leibig \emph{et~al.}, ``Leveraging uncertainty information from deep neural networks for disease detection,'' \emph{Scientific Reports}, vol.~7, no.~1, pp. 1--14, 2017.

\bibitem{liang2018bayesian}
F.~Liang \emph{et~al.}, ``Bayesian neural networks for selection of drug sensitive genes,'' \emph{Journal of the American Statistical Association}, vol. 113, no. 523, pp. 955--972, 2018.

\bibitem{azevedo2020stochasticyolo}
T.~Azevedo \emph{et~al.}, ``Stochastic-yolo: Efficient probabilistic object detection under dataset shifts,'' \emph{arXiv preprint arXiv:2009.02967}, 2020.

\bibitem{neal1993bayesian}
R.~M. Neal, ``Bayesian learning via stochastic dynamics,'' in \emph{Advances in Neural Information Processing Systems (NeurIPS)}, 1993, pp. 475--482.

\bibitem{gal2016dropout}
Y.~Gal and Z.~Ghahramani, ``Dropout as a {Bayesian} approximation: Representing model uncertainty in deep learning,'' in \emph{International Conference on Machine Learning (ICML)}, 2016, pp. 1050--1059.

\bibitem{blundell2015weight}
C.~Blundell \emph{et~al.}, ``Weight uncertainty in neural network,'' in \emph{International Conference on Machine Learning (ICML)}, 2015, pp. 1613--1622.

\bibitem{lakshminarayanan2017simple}
B.~Lakshminarayanan \emph{et~al.}, ``Simple and scalable predictive uncertainty estimation using deep ensembles,'' \emph{Advances in Neural Information Processing Systems}, vol.~30, 2017.

\bibitem{ovadia2019can}
Y.~Ovadia \emph{et~al.}, ``Can you trust your model's uncertainty? evaluating predictive uncertainty under dataset shift,'' \emph{Advances in Neural Information Processing Systems (NeurIPS)}, vol.~32, 2019.

\bibitem{fowers2018configurable}
J.~Fowers \emph{et~al.}, ``A configurable cloud-scale dnn processor for real-time ai,'' in \emph{2018 ACM/IEEE 45th Annual International Symposium on Computer Architecture (ISCA)}.\hskip 1em plus 0.5em minus 0.4em\relax IEEE, 2018, pp. 1--14.

\bibitem{chen2016eyeriss}
Y.-H. Chen \emph{et~al.}, ``Eyeriss: An energy-efficient reconfigurable accelerator for deep convolutional neural networks,'' \emph{IEEE Journal of Solid-State Circuits}, vol.~52, no.~1, pp. 127--138, 2016.

\bibitem{fan2022adaptable}
H.~Fan \emph{et~al.}, ``Adaptable butterfly accelerator for attention-based {NNs} via hardware and algorithm co-design,'' in \emph{MICRO-55: 55th Annual IEEE/ACM International Symposium on Microarchitecture}, 2022.

\bibitem{zhang2018caffeine}
C.~Zhang \emph{et~al.}, ``Caffeine: Toward uniformed representation and acceleration for deep convolutional neural networks,'' \emph{IEEE Transactions on Computer-Aided Design of Integrated Circuits and Systems}, vol.~38, no.~11, pp. 2072--2085, 2018.

\bibitem{fahim2021hls4ml}
F.~Fahim \emph{et~al.}, ``hls4ml: An open-source codesign workflow to empower scientific low-power machine learning devices,'' \emph{arXiv preprint arXiv:2103.05579}, 2021.

\bibitem{krizhevsky2017imagenet}
A.~Krizhevsky \emph{et~al.}, ``Imagenet classification with deep convolutional neural networks,'' \emph{Communications of the ACM}, vol.~60, no.~6, pp. 84--90, 2017.

\bibitem{he2016deep}
K.~He \emph{et~al.}, ``Deep residual learning for image recognition,'' in \emph{{Proceedings of the IEEE Conference on Computer Vision and Pattern Recognition (CVPR)}}, 2016, pp. 770--778.

\bibitem{hochreiter1997long}
S.~Hochreiter and J.~Schmidhuber, ``Long short-term memory,'' \emph{Neural Computation}, vol.~9, no.~8, pp. 1735--1780, 1997.

\bibitem{durasov2021masksembles}
N.~Durasov \emph{et~al.}, ``Masksembles for uncertainty estimation,'' in \emph{Proceedings of the IEEE Conference on Computer Vision and Pattern Recognition (CVPR)}, 2021, pp. 13\,539--13\,548.

\bibitem{ma2017optimizing}
Y.~Ma \emph{et~al.}, ``Optimizing loop operation and dataflow in fpga acceleration of deep convolutional neural networks,'' in \emph{Proceedings of the 2017 ACM/SIGDA International Symposium on Field-Programmable Gate Arrays}, 2017, pp. 45--54.

\bibitem{jospin2022hands}
L.~V. Jospin \emph{et~al.}, ``Hands-on bayesian neural networks—a tutorial for deep learning users,'' \emph{IEEE Computational Intelligence Magazine}, vol.~17, no.~2, pp. 29--48, 2022.

\bibitem{neal2011mcmc}
R.~M. Neal \emph{et~al.}, ``Mcmc using hamiltonian dynamics,'' \emph{Handbook of Markov Chain Monte Carlo}, vol.~2, no.~11, p.~2, 2011.

\bibitem{welling2011bayesian}
M.~Welling and Y.~W. Teh, ``Bayesian learning via stochastic gradient langevin dynamics,'' in \emph{International Conference on Machine Learning (ICML)}, 2011, pp. 681--688.

\bibitem{srivastava2014dropout}
N.~Srivastava \emph{et~al.}, ``Dropout: a simple way to prevent neural networks from overfitting,'' \emph{The journal of machine learning research}, vol.~15, no.~1, pp. 1929--1958, 2014.

\bibitem{laskaridis2021adaptive}
S.~Laskaridis \emph{et~al.}, ``Adaptive inference through early-exit networks: Design, challenges and directions,'' in \emph{Proceedings of the 5th International Workshop on Embedded and Mobile Deep Learning}, 2021, pp. 1--6.

\bibitem{msdnet}
G.~Huang \emph{et~al.}, ``Multi-scale dense networks for resource efficient image classification,'' in \emph{International Conference on Learning Representations (ICLR)}, 2018.

\bibitem{bidistillation}
H.~Lee and J.-S. Lee, ``Students are the best teacher: Exit-ensemble distillation with multi-exits,'' \emph{arXiv preprint arXiv:2104.00299}, 2021.

\bibitem{sdn}
Y.~Kaya \emph{et~al.}, ``Shallow-deep networks: Understanding and mitigating network overthinking,'' in \emph{Proceedings of the 36th International Conference on Machine Learning}, vol.~97.\hskip 1em plus 0.5em minus 0.4em\relax PMLR, 2019, pp. 3301--3310.

\bibitem{fan2022optimizing}
H.~Fan, C.~Guo, and W.~Luk, ``Optimizing quantum circuit placement via machine learning,'' in \emph{2022 59th ACM/IEEE Design Automation Conference (DAC)}, 2022, pp. 19--24.

\bibitem{wan2021shift}
Q.~Wan \emph{et~al.}, ``{Shift-BNN}: Highly-efficient probabilistic bayesian neural network training via memory-friendly pattern retrieving,'' in \emph{2021 54th Annual IEEE/ACM International Symposium on Microarchitecture (MICRO)}, 2021, pp. 885--897.

\bibitem{awano2020bynqnet}
H.~Awano and M.~Hashimoto, ``{BYNQNET}: Bayesian neural network with quadratic activations for sampling-free uncertainty estimation on {FPGA},'' in \emph{2020 Design, Automation \& Test in Europe Conference \& Exhibition (DATE)}.\hskip 1em plus 0.5em minus 0.4em\relax IEEE, 2020, pp. 1402--1407.

\bibitem{fan2022fpga}
H.~Fan \emph{et~al.}, ``{FPGA}-based acceleration for bayesian convolutional neural networks,'' \emph{IEEE Transactions on Computer-Aided Design of Integrated Circuits and Systems}, vol.~41, no.~12, pp. 5343--5356, 2022.

\bibitem{ferianc2021optimizing}
M.~Ferianc \emph{et~al.}, ``Optimizing bayesian recurrent neural networks on an fpga-based accelerator,'' in \emph{2021 International Conference on Field-Programmable Technology (ICFPT)}.\hskip 1em plus 0.5em minus 0.4em\relax IEEE, 2021, pp. 1--10.

\bibitem{cai2018vibnn}
R.~Cai \emph{et~al.}, ``{VIBNN}: Hardware acceleration of bayesian neural networks,'' \emph{ACM SIGPLAN Notices}, vol.~53, no.~2, pp. 476--488, 2018.

\bibitem{fan2022accelerating}
H.~Fan \emph{et~al.}, ``Accelerating {Bayesian} neural networks via algorithmic and hardware optimizations,'' \emph{IEEE Transactions on Parallel and Distributed Systems}, vol.~33, no.~12, pp. 3387--3399, 2022.

\bibitem{ferianc2021effects}
M.~Ferianc \emph{et~al.}, ``On the effects of quantisation on model uncertainty in bayesian neural networks,'' in \emph{Uncertainty in Artificial Intelligence}.\hskip 1em plus 0.5em minus 0.4em\relax PMLR, 2021, pp. 929--938.

\bibitem{bayessegnet}
A.~Kendall \emph{et~al.}, ``{Bayesian Segnet}: Model uncertainty in deep convolutional encoder-decoder architectures for scene understanding,'' \emph{arXiv preprint arXiv:1511.02680}, 2015.

\bibitem{kristiadi2020being}
A.~Kristiadi \emph{et~al.}, ``Being bayesian, even just a bit, fixes overconfidence in relu networks,'' \emph{arXiv preprint arXiv:2002.10118}, 2020.

\bibitem{kendall2015bayesian}
A.~Kendall \emph{et~al.}, ``Bayesian segnet: Model uncertainty in deep convolutional encoder-decoder architectures for scene understanding,'' \emph{arXiv preprint arXiv:1511.02680}, 2015.

\bibitem{ee}
L.~Qendro \emph{et~al.}, ``Early exit ensembles for uncertainty quantification,'' in \emph{Proceedings of Machine Learning for Health}, ser. Proceedings of Machine Learning Research, vol. 158.\hskip 1em plus 0.5em minus 0.4em\relax PMLR, 2021, pp. 181--195.

\bibitem{wan2020fast}
Q.~Wan \emph{et~al.}, ``{Fast-BCNN}: Massive neuron skipping in {Bayesian} convolutional neural networks,'' in \emph{2020 53rd Annual IEEE/ACM International Symposium on Microarchitecture (MICRO)}.\hskip 1em plus 0.5em minus 0.4em\relax IEEE, 2020, pp. 229--240.

\bibitem{pelcat2016design}
M.~Pelcat \emph{et~al.}, ``Design productivity of a high level synthesis compiler versus {HDL},'' in \emph{2016 International Conference on Embedded Computer Systems: Architectures, Modeling and Simulation (SAMOS)}.\hskip 1em plus 0.5em minus 0.4em\relax IEEE, 2016, pp. 140--147.

\end{thebibliography}












\begin{IEEEbiography}[{\includegraphics[width=1in,height=1.25in,clip,keepaspectratio]{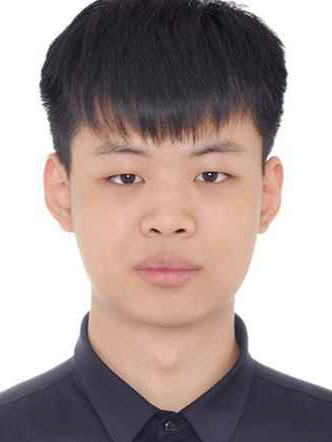}}]{Hao (Mark) Chen}
is a final-year MEng student at Imperial College London. His research interests include machine learning systems, domain-specific languages for embedded systems, and software-hardware co-design.
\end{IEEEbiography}

\begin{IEEEbiography}[{\includegraphics[width=1in,height=1.25in,clip,keepaspectratio]{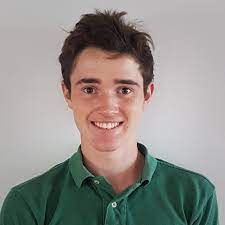}}]{Liam Castelli}
obtained the M.S.c. degree in Artificial Intelligence from the Department of Computing of Imperial College London, London, UK in 2022.
He is currently a Data Engineering Intern at Redica Systems. 
\end{IEEEbiography}

\begin{IEEEbiography}[{\includegraphics[width=1in,height=1.25in,clip,keepaspectratio]{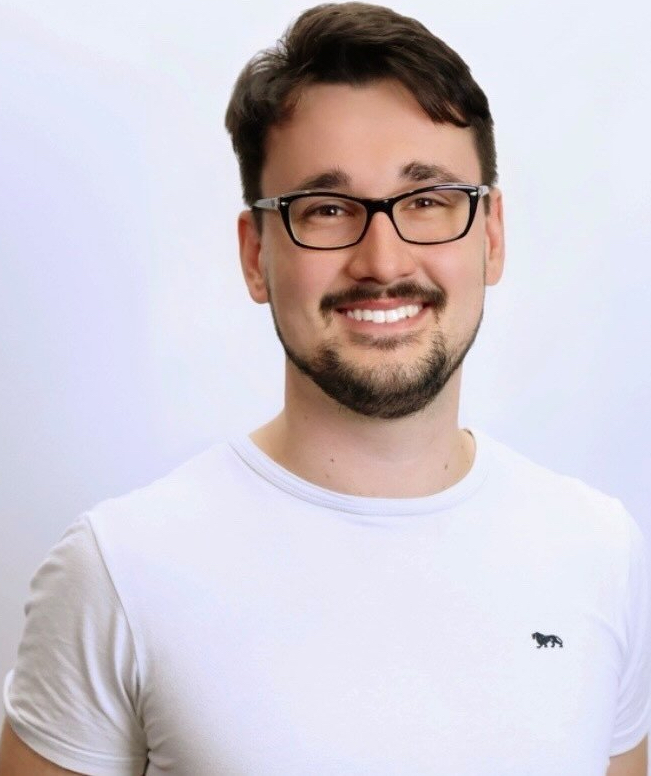}}]{Martin Ferianc}
obtained an MEng in Electronic and Information Engineering from Imperial College London, London, UK in 2015.
He is currently a PhD candidate in the Department of Electronic and Electrical Engineering at University College London.  
His research interests include Bayesian neural networks, deep learning and hardware acceleration of neural networks.
\end{IEEEbiography}

\begin{IEEEbiography}
[{\includegraphics[width=1in,height=1.25in,clip,keepaspectratio]{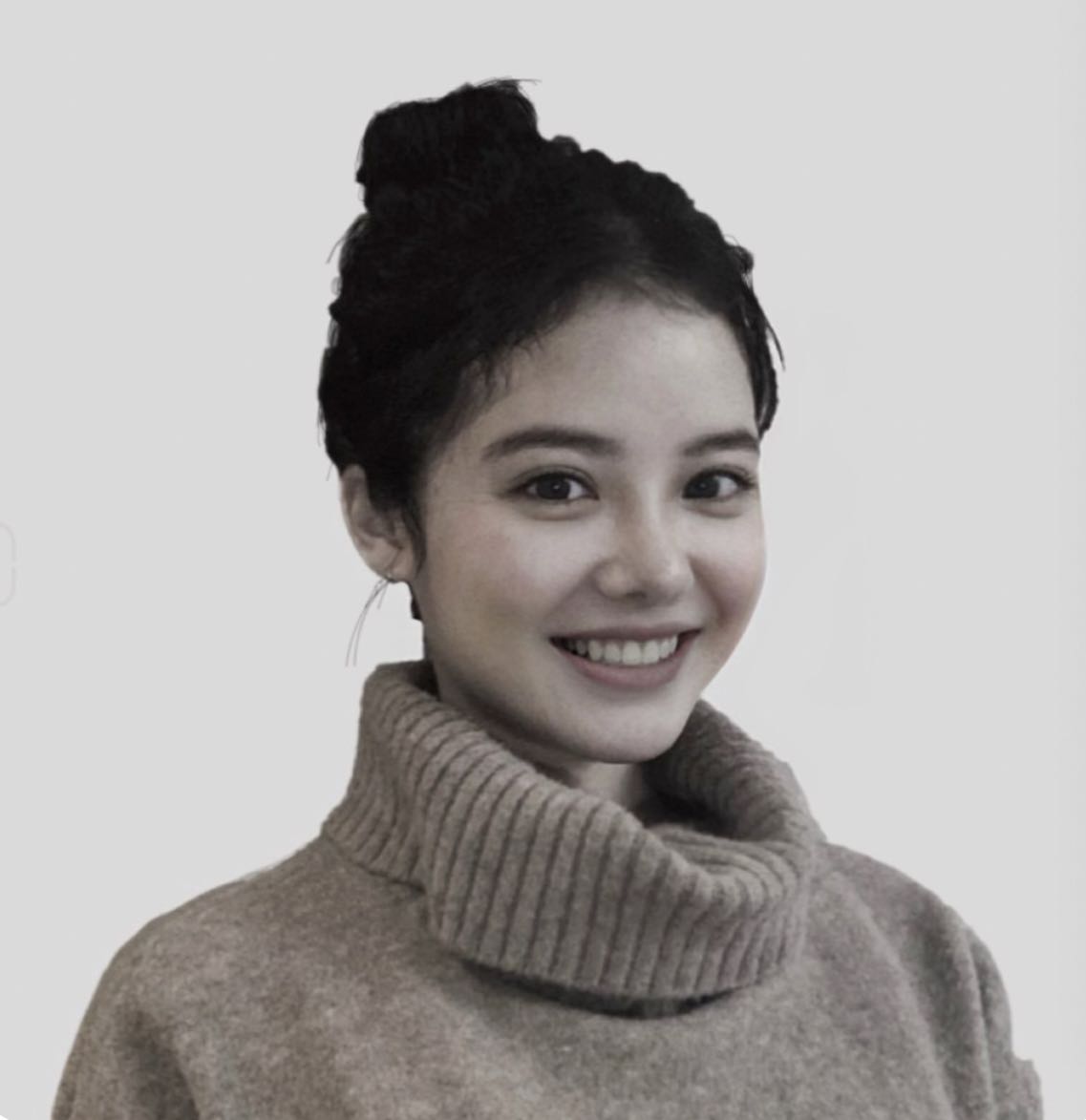}}]
{Hongyu Zhou}
earned her Bachelor’s degree in Industrial Design from Central South University, Changsha, China, in 2019. She received her Master of Research, with a specialization in Design Pathway, from the Royal College of Art, London, UK, in 2021. Currently, she is a Ph.D. candidate at the School of Computer Science, University of Sydney, Sydney, Australia. Her research interests include human-computer interaction, virtual reality, user experience and machine learning.
\end{IEEEbiography}

\begin{IEEEbiography}
[{\includegraphics[width=1in,height=1.25in,clip,keepaspectratio]{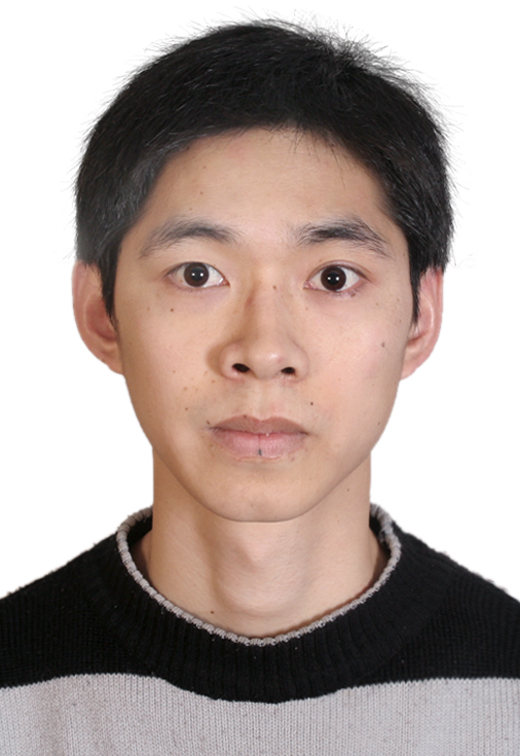}}]
{Shuanglong Liu}
received the B.Sc. and M.Sc. degrees from the Department of Electronic Engineering, Tsinghua University, Beijing, China, in 2010 and 2013 respectively, and Ph.D. degree in Electric Engineering from Imperial
College London, London, U.K, in 2017.
From 2017 to 2020, he was a Research Associate with the Department of Computing, Imperial College London. 
He is currently a Distinguished Professor in the School of Physics and Electronics, Hunan Normal University, Changsha, China. He has published
over 30 research papers in peer-reviewed journals
and international conferences. His current research interests
include reconfigurable and high-performance
computing for Convolutional Neural Networks (CNNs) and statistical inference problems.
\end{IEEEbiography}

\begin{IEEEbiography}[{\includegraphics[width=1in,height=1.25in,clip,keepaspectratio]{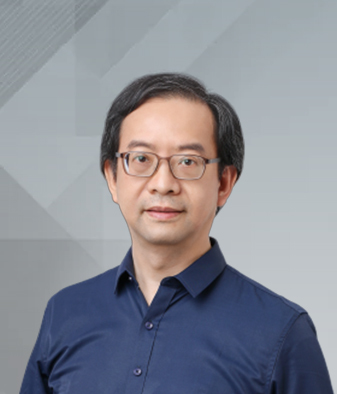}}]{Wayne Luk} (Fellow, IEEE) received the M.A., M.Sc., and D.Phil. degrees in engineering and computing science from Oxford University, Oxford, U.K. He founded and leads the Custom Computing Group, Department of Computing at Imperial College London, where he is Professor of Computer Engineering. He was a Visiting Professor at Stanford University, Stanford, CA, USA. Dr. Luk is a Fellow of the Royal Academy of Engineering and the BCS. He had 15 papers that received awards from international conferences,
and he received a Research Excellence Award from Imperial College London. 
He was a founding Editor-in-Chief of the ACM Transactions on Reconfigurable Technology and Systems, and has been a member of the Steering Committee and Program Committee of various international conferences.
\end{IEEEbiography}

\begin{IEEEbiography}[{\includegraphics[width=1in,height=1.25in,clip,keepaspectratio]{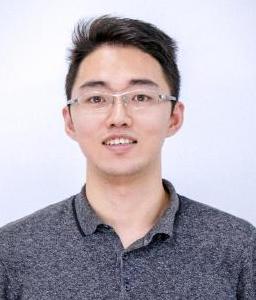}}]{Hongxiang Fan}
received the B.S. degree in electronic
engineering from Tianjin University, Tianjin,
China, in 2017, and the M.Res. and D.Phil. degrees from the Department of Computing, Imperial College London, London, U.K., in 2018 and 2022.
He is currently a research scientist at Samsung AI Cambridge and an affiliated postdoctoral researcher at the University of Cambridge.
His current research focuses on computer architecture, machine learning and quantum computing.
\end{IEEEbiography}

\end{document}